\pdfoutput=1
\documentclass{article}

\usepackage{times}
\usepackage{graphicx} 
\usepackage{subfigure} 

\usepackage{natbib}

\usepackage{algorithm}
\usepackage{algorithmic}

\usepackage{url}
\usepackage{epstopdf}
\usepackage{multirow}
\usepackage{amsmath,amsthm,amssymb}
\usepackage[title]{appendix}

\usepackage{hyperref}

\usepackage{hhline}

\usepackage[accepted]{icml2016}

\icmltitlerunning{Supervised and Semi-supervised Text Categorization
  using LSTM for Region Embeddings}

\begin{document} 

\twocolumn[
\icmltitle{Supervised and Semi-Supervised Text Categorization\\using
  LSTM for Region Embeddings}
  
\icmlauthor{Rie Johnson}{riejohnson@gmail.com}
\icmladdress{RJ Research Consulting, Tarrytown NY, USA}
\icmlauthor{Tong Zhang}{tongzhang@baidu.com}
\icmladdress{Big Data Lab, Baidu Inc, Beijing, China}

\icmlkeywords{keywords}

\vskip 0.3in
]

\begin{abstract} 
One-hot CNN (convolutional neural network) has been shown to be 
effective for text categorization 
\cite{JZ15a,JZ15b}. 
We view it as a special case of a general framework which jointly trains a linear model 
with a non-linear feature generator consisting of `{\em text region embedding + pooling}'.  
Under this framework, 
we explore a more sophisticated region embedding
method using {\em Long Short-Term Memory (LSTM)}.  
LSTM can embed text regions of variable (and possibly large) sizes, whereas the region size needs to be fixed 
in a CNN. 
We seek effective and efficient use of LSTM for this purpose in the supervised and semi-supervised settings. 
The best results were obtained by combining region embeddings in the form of LSTM and convolution layers
trained on unlabeled data.
The results indicate that on this task, 
embeddings of text regions, which can convey complex concepts, 
are more useful than embeddings of single words in isolation. 
We report performances exceeding the previous best results on four benchmark datasets.

\end{abstract} 

\graphicspath{{figures/}} 

\newtheorem{assumption}{Assumption}
\newtheorem{definition}{Definition}
\newtheorem{theorem}{Theorem}

\newcommand{\Xone}{X_1} 
\newcommand{\Xtwo}{X_2}
\newcommand{\cXone}{{\cX_1}}
\newcommand{\cXtwo}{{\cX_2}}
\newcommand{\Xonetwo}{(\Xone,\Xtwo)}

\newcommand{\bowone}{{bow1}}
\newcommand{\bowtwo}{{bow2}}
\newcommand{\bowthree}{{bow3}}
\newcommand{\bongram}{bag-of-$n$-gram}
\newcommand{\bongrams}{bag-of-$n$-grams}

\newcommand{\cX}{{\mathcal X}}
\newcommand{\cY}{{\mathcal Y}}
\newcommand{\cH}{{\mathcal H}}
\newcommand{\bw}{{\mathbf w}}
\newcommand{\bx}{{\mathbf x}}
\newcommand{\bb}{{\mathbf b}}
\newcommand{\bz}{{\mathbf z}}
\newcommand{\by}{{\mathbf y}}
\newcommand{\bp}{{\mathbf p}}
\newcommand{\bv}{{\mathbf v}}

\newcommand{\psz}{p}
\newcommand{\doc}{D}
\newcommand{\voc}{V}
\newcommand{\vsz}{|\voc|}
\newcommand{\wrd}{w}
\newcommand{\dsz}{|\doc|}

\newcommand{\cnn}{CNN}
\newcommand{\cnns}{CNNs}

\newcommand{\ip}[2]{{#1} \cdot {#2}} 
\newcommand{\Ip}[2]{{#1} \cdot {#2}} 
\newcommand{\bias}{b}
\newcommand{\wei}{\bw}
\newcommand{\Wei}{{\mathbf W}}
\newcommand{\Bias}{{\mathbf b}}
\newcommand{\Vei}{{\mathbf V}}

\newcommand{\region}{{\mathbf r}} 
\newcommand{\iL}{\ell} 
\newcommand{\nL}{L}
\newcommand{\ind}{{\cal I}}

\newcommand{\myre}{\mathbb{R}}
\newcommand{\rdim}{q} 
\newcommand{\nodes}{m} 

\newcommand{\tvEmb} {tv-embedding}
\newcommand{\tvEmbs} {tv-embeddings}
\newcommand{\TvEmb} {Tv-embedding}
\newcommand{\TvEmbs} {Tv-embeddings}
\newcommand{\tvEmbd}{tv-embedded}
\newcommand{\tvEmbAbb} {tv-embed.}
\newcommand{\tvEmbUns}  {unsup-tv.}
\newcommand{\tvEmbPar}  {parsup-tv.}
\newcommand{\tvEmbUnsN} {unsup3-tv.}

\newcommand{\bow}{{bow}}

\newcommand{\lstm}{LSTM}
\newcommand{\lstms}{LSTMs}

\newcommand{\ohCnn}{oh-CNN}
\newcommand{\wvLstm}{wv-LSTM}
\newcommand{\wvLstmp}{wv-LSTMp}
\newcommand{\wvLstms}{wv-LSTMs}
\newcommand{\wvBiLstm}{wv-2LSTMp}
\newcommand{\ohLstm}{oh-LSTMp}
\newcommand{\ohBiLstm}{oh-2LSTMp}

\newcommand{\justWx}{{\mathbf W}} 
\newcommand{\justWh}{{\mathbf U}} 
\newcommand{\justbbb}{{\mathbf b}}
\newcommand{\Wx}[1]{\justWx^{(#1)}} 
\newcommand{\Wh}[1]{\justWh^{(#1)}} 
\newcommand{\bbb}[1]{\justbbb^{(#1)}}
\newcommand{\anyWx}{\Wx{\cdot}} 
\newcommand{\anyWh}{\Wh{\cdot}} 
\newcommand{\anybbb}{\bbb{\cdot}}
\newcommand{\Wxi}{\Wx{i}}
\newcommand{\Whi}{\Wh{i}}
\newcommand{\xx}{{\mathbf x}}
\newcommand{\hh}{{\mathbf h}}
\newcommand{\ii}{{\mathbf i}}
\newcommand{\ff}{{\mathbf f}}
\newcommand{\uu}{{\mathbf u}}
\newcommand{\oo}{{\mathbf o}}
\newcommand{\cc}{{\mathbf c}}
\newcommand{\Wtv}[1]{{\widetilde{\mathbf W}^{(#1)}}} 
\newcommand{\xtv}{{\widetilde{\xx}}}

\newcommand{\cnnEmbAbb}{C$_{\rm tv}$embed.}
\newcommand{\lstmEmbAbb}{L$_{\rm tv}$embed.}
\newcommand{\cnnEmb}{C$_{\rm tv}$embedding}
\newcommand{\lstmEmb}{L$_{\rm tv}$embedding}
\newcommand{\cnnEmbs}{\cnnEmb s}
\newcommand{\lstmEmbs}{\lstmEmb s}

\newcommand{\cmt}[1]{{\em #1}\\}

\newcommand{\tightDisplayBegin}[1]{\begingroup \setlength{\belowdisplayskip}{#1} \setlength{\belowdisplayshortskip}{#1} \setlength{\abovedisplayskip}{#1} \setlength{\abovedisplayshortskip}{#1}}
\newcommand{\tightDisplayEnd}{\endgroup}

\newcommand{\tightArrayBegin}[1]{\begingroup \renewcommand\arraystretch{#1}}
\newcommand{\tightArrayEnd}{\endgroup}

\makeatletter
\newcommand\tightpara{\@startsection{paragraph}{4}{\z@}{0.5ex plus
   0ex minus 0.2ex}{-1em}{\normalsize\bf}}   
\makeatother 

\newcommand{\mypara}{\tightpara}

\newcommand{\JZa}{JZ15a}
\newcommand{\JZb}{JZ15b}
\newcommand{\JZab}{JZ15}
\newcommand{\JZaq}{[JZ15a]}
\newcommand{\JZbq}{[JZ15b]}
\newcommand{\DLa}{DL15}
\newcommand{\DLaq}{[DL15]}

\begin{figure*}[t]
\centering
\hfill
\begin{minipage}[b]{0.31\linewidth}
\includegraphics[width=\linewidth]{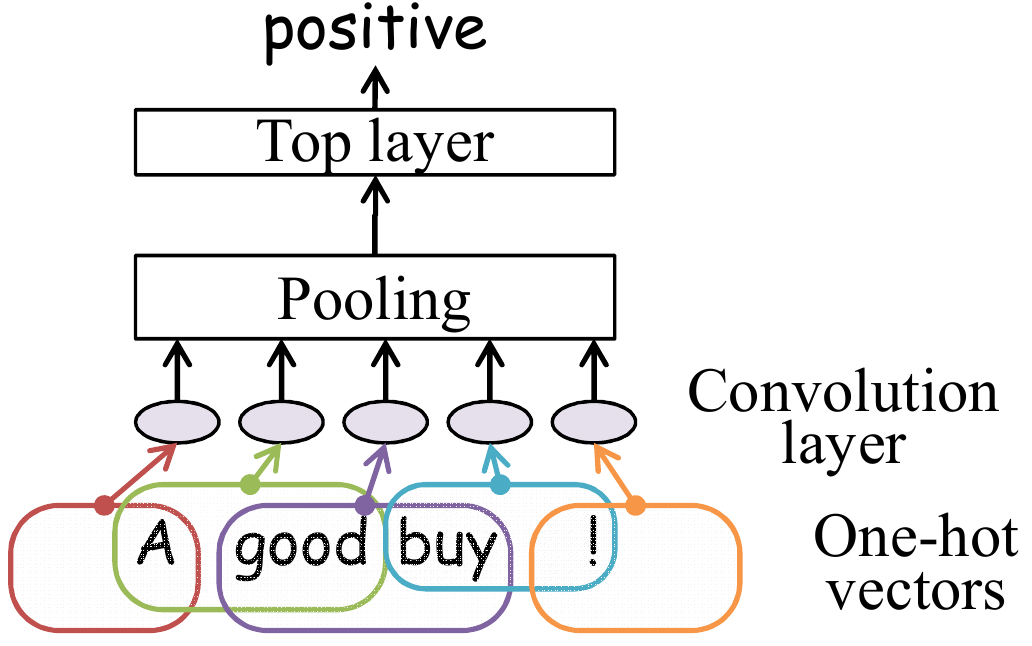}
\vskip -0.1in
\caption{\label{fig:cnn}
One-hot \cnn\ ({\em \ohCnn}) [\JZa] 
}
\end{minipage}
\hfill
\begin{minipage}[b]{0.23\linewidth}
\includegraphics[width=\linewidth]{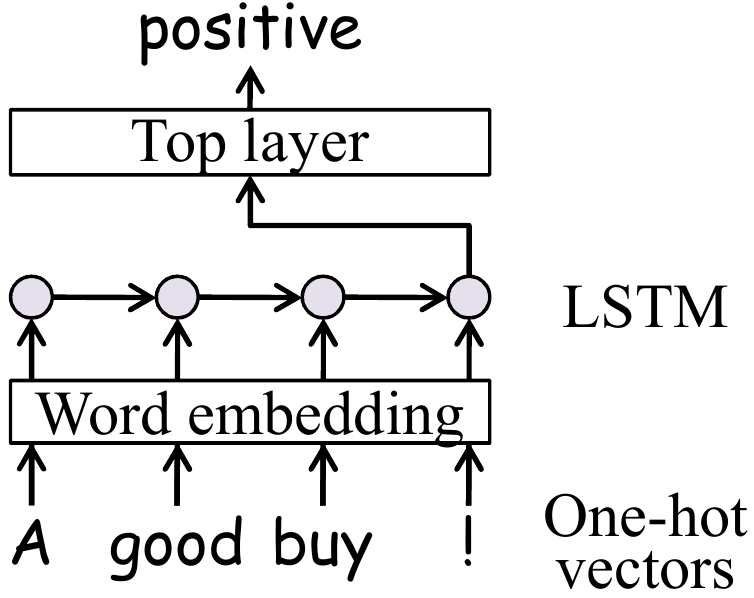}
\vskip -0.1in
\caption{\label{fig:lstmstd} 
Word vector \lstm\ ({\em \wvLstm}) as in \DLaq. 
}
\end{minipage}
\hfill
\begin{minipage}[b]{0.23\linewidth}
\includegraphics[width=\linewidth]{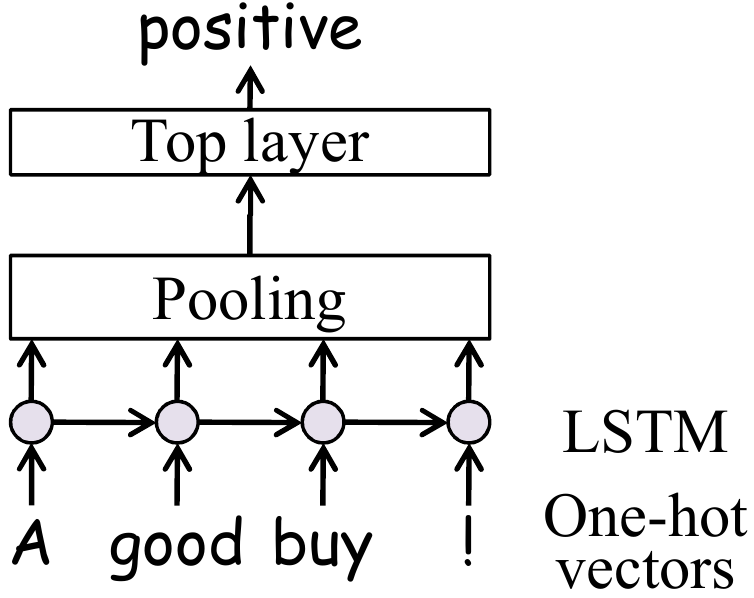}
\vskip -0.1in
\caption{\label{fig:lstmpool} 
One-hot \lstm\ with pooling ({\em \ohLstm}). 
}
\end{minipage}
\hfill
\hfill
\end{figure*}

\section{Introduction} 
\label{sec:intro}

Text categorization is the task of assigning labels to documents written in 
a natural language, and it has numerous real-world applications including sentiment analysis 
as well as traditional topic assignment tasks.  
The state-of-the art methods for text categorization had long been linear predictors 
(e.g., SVM with a linear kernel) with either bag-of-word or bag-of-$n$-gram vectors 
(hereafter {\em \bow}) as input, 
e.g., \cite{J98,LYRL04}. 
%
This, however, has changed recently.  
Non-linear methods that can make effective use of word order 
have been shown to produce more accurate predictors than the traditional \bow-based linear models, e.g., 
\cite{DL15,ZZC15}.  
In particular, let us first focus on {\em one-hot CNN} which we proposed in \JZab\ \cite{JZ15a,JZ15b}.

A {\em convolutional neural network (CNN)} 
\cite{LeCun+etal98} is a feedforward neural network with convolution layers interleaved with pooling layers, 
originally developed for image processing.  
In its convolution layer, a small region of data (e.g., a small square of image) at every location 
is converted to a low-dimensional vector with information relevant to the task being preserved, 
which we loosely term `{\em embedding}'.  
The {\em embedding function} is shared among all the locations, 
so that useful features can be detected irrespective of their locations.  
%
In its simplest form, one-hot \cnn\ works as follows. 
A document is represented as a sequence of {\em one-hot vectors} 
(each of which indicates a word by the position of a 1); 
a convolution layer converts small regions of the document (e.g., ``I love it'') to low-dimensional vectors at every location 
({\em embedding of text regions}); 
a pooling layer aggregates the region embedding results to a document vector
by taking component-wise maximum or average; and 
the top layer classifies a document vector with a linear model (Figure \ref{fig:cnn}). 
The one-hot \cnn\ and its semi-supervised extension were shown to be
superior to a number of previous methods.

In this work, we consider a more general framework (subsuming one-hot \cnn) which 
{\em jointly trains a feature generator and a linear model}, where the 
feature generator consists of `{\em region embedding + pooling}'. 
The specific region embedding function 
of one-hot \cnn\ 
takes the simple form
\newcommand{\cnnAct}[1]{\max(0,#1)}
\begin{align}
\bv(\bx_\iL) = \cnnAct{\Wei \bx_\iL + \bb}~, \label{eq:cnn}
\end{align}
where $\bx_\iL$ is a concatenation of one-hot vectors (therefore, `one-hot' in the name) 
of the words in the $\iL$-th region (of a fixed size), 
and the weight matrix $\Wei$ and the bias vector $\bb$ need to be trained.  
It is simple and fast to compute, and considering its simplicity,
the method works surprisingly well if the region size is appropriately set.  
However, there are also potential shortcomings.  
The region size must be fixed, which may not be optimal as the size of 
relevant regions may vary.  Practically, the region size cannot be very large 
as the number of parameters to be learned (components of $\Wei$) depends on it. 
\JZab\ proposed variations to alleviate these issues.  For example, a bow-input variation allows 
$\bx_\iL$ above to be a bow vector of the region.  This enables a larger region, but 
at the expense of losing word order in the region and so its use 
may be limited.  

In this work, we build on the general framework of `region embedding + pooling' 
and explore a more sophisticated region embedding via {\em Long Short-Term Memory (LSTM)}, 
seeking to overcome the shortcomings above, 
in the supervised and semi-supervised settings. 
\lstm\ \cite{HS97} is a recurrent neural network.  In its typical applications to text, 
an \lstm\ takes words in a sequence one by one; i.e., 
at time $t$, it takes as input the $t$-th word and the output from time $t-1$.  
Therefore, the output from each time step can be regarded as the embedding of the sequence of words 
that have been seen so far (or a relevant part of it). 
It is designed to enable learning of dependencies over larger time lags 
than feasible with traditional recurrent networks. 
That is, an \lstm\ can be used to embed text regions of variable (and possibly large) sizes. 

We pursue the best use of \lstm\ for our purpose, and then compare the resulting model 
with the previous best methods including one-hot \cnn\ and previous \lstm.  
Our strategy is to {\em simplify the model} as much as possible, 
including elimination of a word embedding layer routinely used to produce input to \lstm. 
Our findings are three-fold.  
First, in the supervised setting, 
our simplification strategy 
leads to higher accuracy and faster training than previous \lstm. 
Second, 
accuracy can be further improved by training 
LSTMs on unlabeled data 
for learning useful {\em region embeddings} 
and using them to produce {\em additional input}. 
Third, 
both our \lstm\ models and one-hot \cnn\ 
strongly outperform other methods including 
previous \lstm.  The best results are obtained by combining the two types of region embeddings 
({\em \lstm\ embeddings} and {\em \cnn\ embeddings}) trained on unlabeled data,  
indicating that their strengths are complementary. 
Overall, our results show that for text categorization, 
embeddings of text regions, which can convey higher-level concepts than single words in isolation, 
are useful, 
and that useful region embeddings can be learned 
without going through word embedding learning. 
%
We report performances exceeding the previous best results on four benchmark datasets. 
Our code and experimental details are available 
at http://riejohnson.com/cnn\_download.html. 

\subsection{Preliminary} 

On text, \lstm\ has been used for labeling or generating words.  
It has been also used for representing short sentences 
mostly for sentiment analysis, and some of them rely on syntactic parse trees; 
see e.g., \cite{ZSG15,TQL15,TSM15,LZ15}. 
Unlike these studies, 
this work as well as \JZab\ 
focuses on classifying general full-length documents without any special linguistic knowledge. 
Similarly, 
\DLa\ \cite{DL15} applied \lstm\ to categorizing general full-length documents. 
Therefore, 
our empirical comparisons will focus on \DLa\ and \JZab, both of which reported 
new state of the art results. 
Let us first introduce the general \lstm\ formulation, 
and then briefly describe \DLa's model as it illustrates the challenges in using \lstms\ for this task.

\mypara{\lstm} 
While several variations exist, 
we base our work on the following \lstm\ formulation, 
which was used in, e.g., \cite{ZI14} 
\newcommand{\squash}{\sigma}
\begin{align*}
\ii_t &= \squash(\Wx{i}\xx_t + \Wh{i}\hh_{t-1} + \bbb{i} )~, \\
\oo_t &= \squash(\Wx{o}\xx_t + \Wh{o}\hh_{t-1} + \bbb{o} )~, \\
\ff_t &= \squash(\Wx{f}\xx_t + \Wh{f}\hh_{t-1} + \bbb{f} )~, \\
\uu_t &= \tanh(\Wx{u}\xx_t + \Wh{u}\hh_{t-1} + \bbb{u} )~, \\
\cc_t &= \ii_t \odot \uu_t + \ff_t \odot \cc_{t-1}~, \\
\hh_t &= \oo_t \odot \tanh(\cc_t)~,
\end{align*}
where $\odot$ denotes element-wise multiplication and 
$\squash$ is an element-wise squash function to make the gating values in $[0,1]$.  
We fix $\squash$ to sigmoid.  $\xx_t \in \myre^d$ is the input from the lower layer 
at time step $t$, where $d$ would be, for example, 
size of vocabulary if the input was a one-hot vector representing a word, 
or the dimensionality of word vector if the lower layer was a word embedding layer. 
With $q$ LSTM units, the dimensionality of the weight matrices and bias vectors, which need to be trained, are 
$\anyWx{\cdot} \in \myre^{q\times d}$, $\anyWh \in \myre^{q\times q}$, and $\anybbb \in \myre^q$ for all types ($i,o,f,u$). 
The centerpiece of \lstm\ is 
the {\em memory cells} 
$\cc_t$, 
designed to counteract the risk of vanishing/exploding gradients, 
thus enabling learning of dependencies over larger time lags 
than feasible with traditional recurrent networks. 
The {\em forget gate} $\ff_t$ \cite{GSC00} is for 
resetting the memory cells.  
The {\em input gate} $\ii_t$ and {\em output gate} $\oo_t$ control 
the input and output of the memory cells.  

\mypara{Word-vector \lstm\ (\wvLstm) \DLaq\ } 
\DLa's application of \lstm\ to text categorization is straightforward.  
As illustrated in Figure \ref{fig:lstmstd}, 
for each document, 
the output of the \lstm\ layer is the output of the last time step (corresponding to the last word of the document), 
which represents the whole document (document embedding). 
Like many other studies of \lstm\ on text, words are first converted to 
low-dimensional dense word vectors via a word embedding layer; therefore, 
we call it {\em word-vector \lstm} or {\em \wvLstm}. 
\DLa\ observed that 
\wvLstm\ underperformed linear predictors and its training was unstable.  
This was attributed to the fact that documents are long. 

In addition, we found that training and testing of \wvLstm\ is time/resource consuming. 
To put it into perspective, using a GPU, one epoch of \wvLstm\ training 
takes nearly 20 times longer than that of one-hot \cnn\ training even though it achieves poorer accuracy 
(the first two rows of Table \ref{tab:time}). 
This is due to the sequential nature of \lstm, i.e., computation at time $t$ requires the output of time $t-1$, 
whereas modern computation depends on parallelization 
for speed-up.  
Documents in a mini-batch can be processed in parallel, but the variability of document lengths 
reduces the degree of parallelization\footnote{
  \cite{SVL14} suggested making each mini-batch consist of sequences of similar lengths, but we found that on our tasks 
  this strategy slows down convergence presumably by hampering the stochastic nature of SGD. 
}. 

It was shown in \DLa\ that training becomes stable and accuracy improves drastically 
when \lstm\ and the word embedding layer are jointly {\em pre-trained} with either the language model learning 
objective (predicting the next word) or autoencoder objective (memorizing the document).  

\section{Supervised \lstm\ for text categorization}
\label{sec:sup}

Within the framework of `region embedding + pooling' for text categorization, 
we seek effective and efficient use of \lstm\ 
as an alternative region embedding method. 
This section focuses on an end-to-end supervised setting so that 
there is no additional data (e.g., unlabeled data) or additional algorithm (e.g., for learning a word embedding). 
Our general strategy is to simplify the model as much as possible. 
We start with 
elimination of the word embedding layer  
so that one-hot vectors are directly fed to \lstm, 
which we call {\em one-hot \lstm} in short.  
%

\subsection{Elimination of the word embedding layer} 
\label{sec:onehot}
\newcommand{\wordW}{{\mathbf V}}
%
Facts: A word embedding is a linear operation that can be written as $\wordW\xx_t$ 
with $\xx_t$ being a one-hot vector and columns of $\wordW$ being word vectors.  
Therefore, 
by replacing the \lstm\ weights $\anyWx$ with $\anyWx\wordW$ and removing the word embedding layer, 
a word-vector \lstm\ can be turned into a one-hot \lstm\ without changing the model behavior.  
Thus, word-vector \lstm\ is not more expressive than one-hot \lstm; rather, 
a merit, if any, 
of 
training with a word embedding layer would be through imposing restrictions
(e.g., a low-rank $\wordW$ makes a less expressive model) 
to achieve good prior/regularization effects. 

In the end-to-end supervised setting, 
a word embedding matrix $\wordW$ 
would need to be initialized randomly and trained as part of the model. 
In the preliminary experiments under our framework, 
we were unable to improve accuracy over one-hot \lstm\ 
by inclusion of such a randomly initialized word embedding layer; i.e., 
random vectors failed to provide good prior effects. 
Instead, demerits 
were evident -- 
more meta-parameters to tune, 
poor accuracy with low-dimensional word vectors, and 
slow training/testing with high-dimensional word vectors as they are dense. 

If a word embedding is appropriately pre-trained with unlabeled data, 
its inclusion is a form of semi-supervised learning and could be useful.  
We will show later, however, 
that this type of approach falls behind our approach of learning {\em region embeddings} 
through training one-hot \lstm\ on unlabeled data.   
Altogether, elimination of the word embedding layer was found to be useful; 
thus, we base our work on one-hot \lstm. 

\subsection{More simplifications} 

We introduce four more useful modifications to \wvLstm\ 
that lead to higher accuracy or faster training.

\mypara{Pooling: simplifying sub-problems} 
Our framework of `region embedding + pooling' has a simplification effect as follows.   
In \wvLstm, the sub-problem that \lstm\ needs to solve is to represent the entire document 
by one vector ({\em document embedding}). 
We make this easy by changing it to detecting regions of text (of arbitrary sizes)  
that are relevant 
to the task and representing them by vectors ({\em region embedding}). 
As illustrated in Figure \ref{fig:lstmpool}, we let the \lstm\ layer emit vectors $\hh_t$ at each time step, 
and let pooling aggregate them into a document vector.  
With \wvLstm, \lstm\ has to remember relevant information until it gets to the end of 
the document even if relevant information was observed 10K words away. 
The task of our \lstm\ is easier as it is allowed to forget old things via the forget gate 
and can focus on representing the concepts conveyed by smaller segments such as phrases or sentences.  

A related architecture appears in the Deep Learning Tutorials\footnote{
  {http://deeplearning.net/tutorial/lstm.html}
}
though it uses a word embedding.
Another related work is
%
\cite{LXLZ15}, which combined pooling with non-LSTM recurrent networks and a word embedding. 

\mypara{Chopping for speeding up training} 
In addition to simplifying the sub-problem, pooling has the merit of enabling faster training via {\em chopping}. 
Since we set the goal of \lstm\ to embedding text regions instead of documents, 
it is no longer crucial to go through the document from the beginning to the end sequentially.  
At the time of training, we can {\em chop} each document into segments of a fixed length that is sufficiently long (e.g., 50 or 100) 
and process all the segments in a mini batch in parallel as if these segments were individual documents.  
(Note that this is done only in the \lstm\ layer and pooling is done over the entire document.) 
We perform testing without chopping.  That is, we train \lstm\ with 
approximations of sequences for speed up and test with real sequences for better accuracy.  
There is a risk of chopping important phrases 
(e.g., ``don't $|$ like it''), and this can be easily avoided by having segments slightly overlap.  
However, we found that gains from overlapping segments tend to be small and so our experiments 
reported below were done without overlapping.  

\mypara{Removing the input/output gates} 
We found that when \lstm\ is followed by pooling, the presence of input and output gates typically 
does not improve accuracy, while 
removing them nearly halves the time and memory required for training and testing.  
It is intuitive, in particular, that pooling can make the output gate unnecessary; 
the role of the output gate is to prevent undesirable information from entering the output $\hh_t$, and  
such irrelevant information can be filtered out by max-pooling. 
Without the input and output gates, the \lstm\ formulation can be simplified to: 
\begin{align}
\ff_t &= \squash(\Wx{f}\xx_t + \Wh{f}\hh_{t-1} + \bbb{f} )~, \label{eq:f}\\
\uu_t &= \tanh(\Wx{u}\xx_t + \Wh{u}\hh_{t-1} + \bbb{u} )~, \label{eq:u}\\
\cc_t &= \uu_t + \ff_t \odot \cc_{t-1}~,~~~\hh_t = \tanh(\cc_t)~. \nonumber
\end{align}
This is equivalent to fixing $\ii_t$ and $\oo_t$ to all ones.  
It is in spirit similar to Gated Recurrent Units \cite{Cho+etal14} but simpler, having fewer gates. 

\begin{table}
\begin{center}
\begin{tabular}{|l|r|r|r|}
\hline
   &{\small Chop}&{\small Time}&{\small Error}\\
\hline
1-layer \ohCnn & -- & 18 & 7.64 \\
\hline
\wvLstm\                & --  & 337 & 11.59 \\ 
\hline
\wvLstmp\               & 100 & 110 & 10.90 \\ 
\ohLstm\                & 100 & 88  & 7.72 \\ 
\ohLstm; no i/o gates   & 100 & 48 & 7.68 \\ 
\ohBiLstm; no i/o gates & 50  & 84 & 7.33 \\ 
\hline
\end{tabular}
\vskip -0.05in
\caption{
\label{tab:time}
Training time and error rates of \lstms\ on Elec. 
``Chop'': chopping size. 
``Time'': seconds per epoch for training on Tesla M2070. 
``Error'': classification error rates (\%) on test data. 
``\wvLstmp'': word-vector \lstm\ with pooling. 
``\ohLstm'': one-hot \lstm\ with pooling. 
``\ohBiLstm'': one-hot bidirectional \lstm\ with pooling. 
}
\end{center}
\end{table}

\begin{figure}
\centering
\includegraphics[width=2.42in]{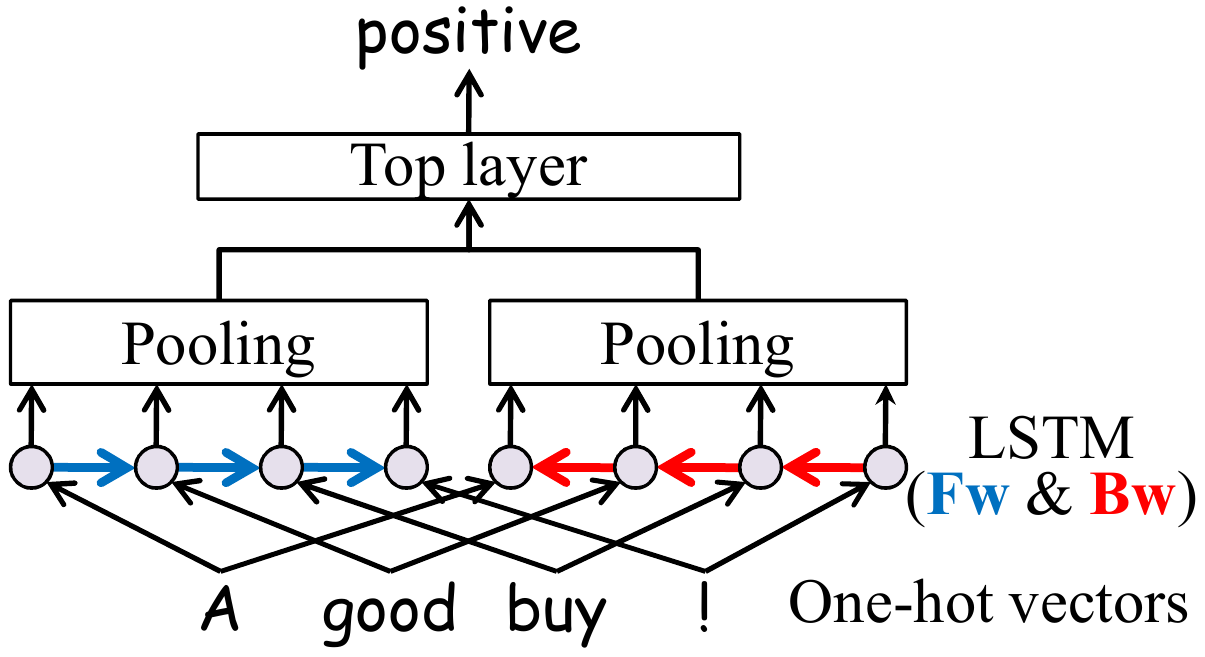}
\vskip -0.15in
\caption{\label{fig:lstmbipool}
{\em \ohBiLstm}: our one-hot bidirectional LSTM with pooling. 
}
\end{figure}
\mypara{Bidirectional \lstm\ for better accuracy} 
The changes from \wvLstm\ 
above substantially reduce the time and memory required for training 
and make it practical to add one more layer of \lstm\ going in the opposite direction 
for accuracy improvement.  
As shown in Figure \ref{fig:lstmbipool}, 
we concatenate the output of a forward \lstm\ (left to right) and 
a backward \lstm\ (right to left), which is referred to as {\em bidirectional \lstm} in the literature. 
The resulting model is a {\em one-hot bidirectional \lstm\ with pooling}, and we abbreviate it 
to {\em \ohBiLstm}.  
Table \ref{tab:time} shows 
how much accuracy and/or training speed can be improved 
by elimination of the word embedding layer, 
pooling, chopping, removing the input/output gates, 
and adding the backward \lstm.  

\begin{table}[t]
\begin{center}
\begin{tabular}{|c|c|c|r|r|c|}
\hline
    & \#train & \#test &  avg & max & \#class \\     
\hline
IMDB & 25,000 & 25,000 &  265&  3K &2\\ 
\hline
Elec & 25,000 & 25,000 &  124&  6K &2\\ 
\hline
RCV1 & 15,564 & 49,838 &  249& 12K &55\\ 
\hline
20NG & 11,293 &  7,528 &  267& 12K &20\\ 
\hline
\end{tabular}
\caption{ \label{tab:data} 
Data.  ``avg''/``max'': the average/maximum length of documents (\#words) of the training/test data. 
IMDB and Elec are for sentiment classification (positive vs. negative) of movie reviews and Amazon 
electronics product reviews, respectively.  RCV1 (second-level topics only) and 20NG are for topic categorization of 
Reuters news articles and newsgroup messages, respectively.  
}
\end{center}
\end{table}

\subsection{Experiments (supervised)} 
\label{sec:supexp}

We used four datasets: 
IMDB, 
Elec, 
RCV1 (second-level topics), 
and 
20-newsgroup (20NG)\footnote{
  http://ana.cachopo.org/datasets-for-single-label-text-categorization
}, to facilitate direct comparison with \JZab\ and \DLa.  
The first three were used in \JZab.  
IMDB and 20NG 
were used in \DLa.  
The datasets are summarized in Table \ref{tab:data}. 

The data was converted to lower-case letters.  
In the neural network experiments, 
vocabulary was reduced to 
the most frequent 30K words of the training data to reduce computational burden; 
square loss was minimized with 
dropout \cite{dropout12} applied to the input to the top layer; 
weights were initialized by the Gaussian distribution with zero mean and 
standard deviation 0.01.  
Optimization was done with SGD with mini-batch size 50 or 100 
with momentum or optionally 
{\em rmsprop} \cite{rmsprop} for acceleration.  

Hyper parameters such as learning rates 
were chosen based on the performance on the development data, 
which was a held-out portion of the training data, and training was redone using all the 
training data with the chosen parameters.

We used the same pooling method as used in \JZab, which parameterizes the number of pooling regions so that 
pooling is done for $k$ non-overlapping regions of equal size, and the resulting $k$ vectors are concatenated 
to make one vector per document.  
The pooling settings chosen based on the performance on the development data are 
the same as \JZa, which are 
max-pooling with $k$=1 on IMDB and Elec and average-pooling with $k$=10 on RCV1; 
on 20NG, max-pooling with $k$=10 was chosen.  

\newcommand{\dl}{}
\newcommand{\jza}{}
\newcommand{\jzb}{}

\begin{table}[h]
\begin{center}
\begin{tabular}{|l|r|r|r|r|}
\hline
          methods         &{\small IMDB}&\multicolumn{1}{|c|}{\small Elec}&{\small RCV1}&{\small 20NG}\\
\hline
      SVM bow & 11.36 & 11.71 & {\em 10.76} & 17.47 \\
SVM 1--3grams &{\em 9.42}&{\em 8.71}&{\em 10.69}&{\em 15.85}\\
\hline
\wvLstm\ {\small \DLaq\ }  & 13.50 & 11.74 & 16.04 & 18.0$~~$\\
 {\bf \ohBiLstm}           &{\bf 8.14}&{\bf 7.33}& 11.17 &{\bf 13.32}\\
\ohCnn\ {\small \JZbq}  & 8.39 & 7.64 & {\bf 9.17}& 13.64\\
 \hline                    
\end{tabular}
\caption{ \label{tab:sup}
Error rates (\%).  Supervised results without any pre-training.  
SVM and \ohCnn\ results on all but 20NG are from \JZa\ and \JZb, 
respectively; 
\wvLstm\ results on IMDB and 20NG are from \DLa; 
all others are new experimental results of this work.  
}
\end{center}
\end{table}


Table \ref{tab:sup} shows the error rates obtained without any additional unlabeled data 
or pre-training of any sort. 
For meaningful comparison, this table 
shows neural networks with comparable dimensionality of embeddings, 
which are one-hot \cnn\ with one convolution layer with 1000 feature maps 
and bidirectional \lstms\ of 500 units each.  
In other words, the convolution layer produces a 1000-dimensional vector at each location, 
and the \lstm\ in each direction emits a 500-dimensional vector at each time step. 
An exception is 
\wvLstm, equipped with 512 \lstm\ units (smaller than 2$\times$500) 
and a word embedding layer of 512 dimensions; 
\DLa\ states that without pre-training, addition of more \lstm\ units broke down training.  
A more complex and larger one-hot \cnn\ will be reviewed later. 

Comparing the two types of \lstm\ in Table \ref{tab:sup}, we see that our one-hot bidirectional \lstm\ with pooling (\ohBiLstm) 
outperforms word-vector \lstm\ (\wvLstm) on all the datasets, confirming the effectiveness of our approach. 
%

Now we review the non-\lstm\ baseline methods. 
The last row of Table \ref{tab:sup} shows the best one-hot \cnn\ results 
within the constraints above.  
They were obtained by {\em bow-\cnn} (whose input to the embedding function (\ref{eq:cnn}) is a bow vector of the region) with region size 20 
on RCV1, and {\em seq-\cnn} (with the regular concatenation input) 
with region size 3 on the others.   
In Table \ref{tab:sup}, 
on three out of the four datasets, \ohBiLstm\ outperforms SVM and the \cnn. 
However, on RCV1, it underperforms both.    
We conjecture that this is because strict word order is not very useful on RCV1. 
This point can also be observed in the SVM and \cnn\ performances. 
Only on RCV1, $n$-gram SVM is no better than bag-of-word SVM, and 
only on RCV1, bow-\cnn\ outperforms seq-\cnn.  
That is, on RCV1, 
bags of words in a window of 20 at every location are more useful than words in strict order. 
This is 
presumably because the former can more easily cover variability of expressions indicative of topics. 
Thus, \lstm, which does not have an ability to put words into bags, loses to bow-\cnn.  

\begin{center}
\begin{tabular}{|l|r|r|r|}
\hline
          methods         &{\small IMDB}&{\small Elec}&{\small 20NG}\\
\hline
 \ohBiLstm, {\small copied from Tab.\ref{tab:sup}}& 8.14 & 7.33 & 13.32 \\
\ohCnn, {\small 2 region sizes \JZaq\ }& 8.04 & 7.48 & 13.55 \\ 
\hline                    
\end{tabular}
\end{center}

\mypara{More on one-hot CNN vs. one-hot LSTM} 
\lstm\ can embed regions of variable (and possibly large) sizes whereas \cnn\ requires the region size to be fixed. 
We attribute to this fact the small improvements of 
\ohBiLstm\ 
over 
\ohCnn\ 
in Table \ref{tab:sup}.  
However, this shortcoming of \cnn\ can be alleviated by having multiple convolution layers 
with distinct region sizes. 
We show in the table above that one-hot \cnns\ with two layers (of 1000 feature maps each) 
with two different region sizes\footnote{
  Region sizes were 2 and 3 for IMDB, 3 and 4 for Elec, and 3 and 20 (bow input) for 20NG. 
}
rival 
\ohBiLstm. 
Although these models are larger than those in Table \ref{tab:sup}, training/testing is still faster than the 
\lstm\ models due to simplicity of the region embeddings.  
By comparison, 
the strength of \lstm\ to embed larger regions appears not to be a big contributor here.  
This may be because the amount of training data is not sufficient enough to learn the relevance of longer word sequences. 
Overall, 
one-hot \cnn\ works surprising well considering its simplicity, and 
this observation motivates the idea of {\em combining the two types of region embeddings}, 
discussed later.  
 
\mypara{Comparison with the previous best results on 20NG} 
The previous best performance on 20NG is 15.3 (not shown in the table) 
of \DLa, obtained by {\em pre-training} 
\wvLstm\ of 1024 units with labeled training data.  
Our \ohBiLstm\ achieved 13.32, which is 2\% better.  
The previous best results on the other datasets use unlabeled data, 
and we will review them with our semi-supervised results. 

\section{Semi-supervised \lstm}
\label{sec:semi}

%
To exploit unlabeled data as an additional resource, 
we use a non-linear extension of {\em two-view feature learning}, 
whose linear version appeared in our earlier work \cite{AZ05jmlr,AZ07}.  
This was used in \JZb\ to learn from unlabeled data a region embedding embodied by a convolution layer. 
In this work we use it to learn a region embedding embodied by a one-hot \lstm.   
Let us start with a brief review of non-linear two-view feature learning.   

\subsection{Two-view embedding (\tvEmb) [\JZb]} 
\label{sec:tvemb}
A rough sketch is as follows. 
Consider two views of the input.  
An embedding is called a {\em \tvEmb} if the embedded view is as good as the original view 
for the purpose of predicting the other view.  
If the two views and the labels (classification targets) are related to one another 
only through some hidden states, then the \tvEmbd\ view is as good as the original view 
for the purpose of classification.  Such an embedding is useful provided that its dimensionality is 
much lower than the original view.  

\JZb\ applied this idea by regarding text regions embedded by the convolution layer  
as one view and their surrounding context as the other view and training a \tvEmb\ (embodied by a convolution layer) 
on unlabeled data.  The obtained \tvEmbs\ were used to produce additional input to 
a supervised region embedding of one-hot \cnn,
resulting in higher accuracy.  

\subsection{Learning \lstm\ \tvEmbs} 

\begin{figure}[h]
\centering
\includegraphics[width=2.5in]{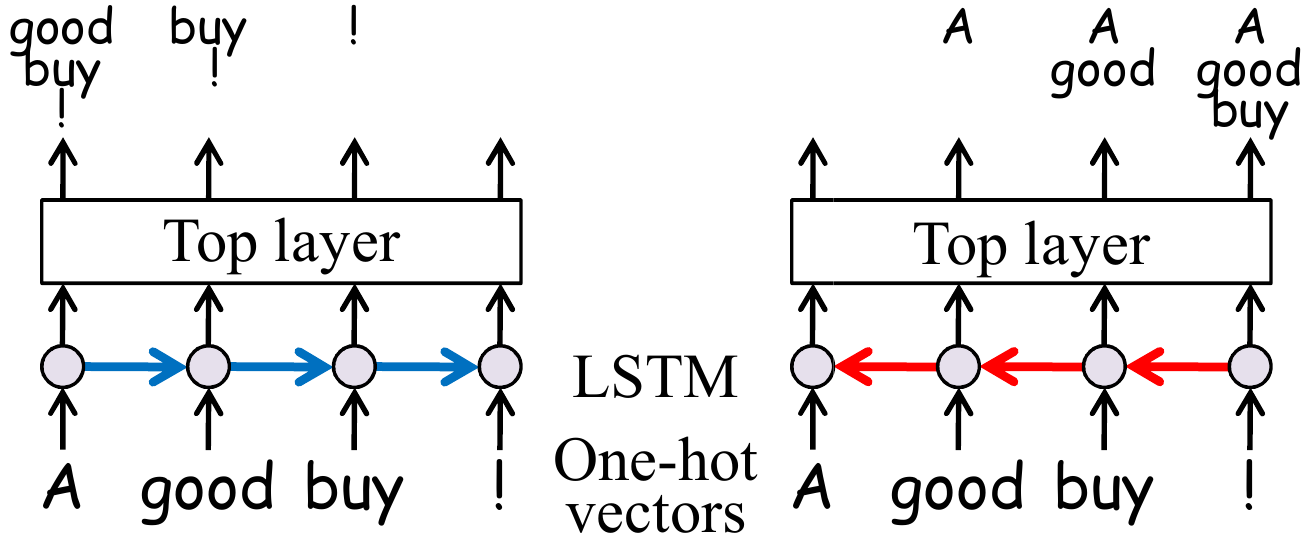}
\vskip -0.15in
\caption{\label{fig:lstmuns}
Training \lstm\ \tvEmbs\ on unlabeled data
}
\end{figure}

\begin{table}[t]
\begin{center}
\begin{small}
\begin{tabular}{|c|rl|l|}   
\hline
IMDB & 75K &(20M words) & Provided \\
\hline
Elec & 200K &(24M words) & Provided \\
\hline
RCV1 & 669K &(183M words) & Sept'96--June'97 \\
\hline
\end{tabular}
\end{small}
\vskip -0.05in
\caption{ \label{tab:unlab} 
Unlabeled data.  See \JZb\ for more details. 
}
\end{center}
\end{table}
In this work
we obtain a \tvEmb\ in the form of \lstm\ from unlabeled data as follows. 
At each time step, we consider the following two views: the words we have already seen in the document (view-1), 
and the next few words (view-2).  The task of \tvEmb\ learning is 
to predict view-2 based on view-1.  We train one-hot \lstms\ in both directions, as in Figure \ref{fig:lstmuns}, 
on unlabeled data.  For this purpose, we use the input and output gates as well as the forget gate 
as we found them to be useful.  

The theory of \tvEmb\ says that 
the region embeddings obtained in this way are useful for the task of interest {\em if}
the two views are related to each other through the concepts relevant to the task. 
To reduce undesirable relations between the views such as syntactic relations, \JZb\ performed 
vocabulary control to remove function words from (and only from) 
the vocabulary of the target view, which we found useful also for \lstm.   

\newcommand{\embIdx}{j}
\newcommand{\embSet}{S}
We use the \tvEmbs\ obtained from unlabeled data to produce {\em additional input} to 
\lstm\ by replacing (\ref{eq:f}) and (\ref{eq:u}) by the following: 
\begin{align*}
\ff_t &= \squash(\Wx{f}\xx_t + \sum_{\embIdx\in\embSet} \Wtv{\embIdx,f}\xtv^{\embIdx}_t + \Wh{f}\hh_{t-1} + \bbb{f} )~, \\ 
\uu_t &= \squash(\Wx{u}\xx_t + \sum_{\embIdx\in\embSet} \Wtv{\embIdx,u}\xtv^{\embIdx}_t + \Wh{u}\hh_{t-1} + \bbb{u} )~.    
\end{align*}
$\xtv^{\embIdx}_t$ is the output of a \tvEmb\ (an \lstm\ trained with unlabeled data)
indexed by $\embIdx$ at time step $t$, and 
$\embSet$ is a set of \tvEmbs\ which contains the two \lstms\ going forward and backward as in Figure \ref{fig:lstmuns}. 
Although it is possible to fine-tune the \tvEmbs\ with labeled data, for simplicity and faster training, 
we fixed them in our experiments. 

\begin{table*}
\begin{center}
\begin{tabular}{c|l|l|r|r|r|}
\cline{2-6} 
&         &\multicolumn{1}{|c|}{Unlabeled data usage} & IMDB & Elec & RCV1 \\
\cline{2-6}       
1 & \wvLstm\ {\small \DLaq\ }& Pre-training                &7.24 & 6.84 & 14.65 \\ 
\cline{2-6} 
2 & \multirow{2}{*}{\wvBiLstm} & 300-dim Google News word2vec    & 8.67 & 7.64 & 10.62 \\ 
\cline{3-6}
3 &                        & 200-dim word2vec scaled & 7.29 & 6.76 & 10.18 \\ 
\cline{2-6} 
4 & {\bf \ohBiLstm}           & $2\times$100-dim \lstm\ \tvEmbAbb\ &{\bf 6.66} &{\bf 6.08} & 9.24 \\
\cline{2-6} 
5 & \ohCnn\ {\small \JZbq\ }  & $1\times$200-dim \cnn\ \tvEmbAbb   & 6.81 & 6.57 &{\bf 7.97} \\
\cline{2-6} 
\end{tabular}
\caption{ \label{tab:semi}
Semi-supervised error rates (\%). 
The \wvLstm\ result on IMDB is from \DLaq; the \ohCnn\ results are from \JZbq; all others are the results of our new experiments.  
}
\end{center}
\end{table*}

\subsection{Combining \lstm\ \tvEmbs\ and \cnn\ \tvEmbs}
\label{sec:combo}

It is easy to see that the set $\embSet$ above can be expanded with {\em any \tvEmbs}, 
not only those in the form of \lstm\ ({\em \lstm\ \tvEmbs}) 
but also with the \tvEmbs\ in the form of convolution layers ({\em \cnn\ \tvEmbs}) 
such as those obtained in \JZb. 
Similarly, 
it is possible to use \lstm\ \tvEmbs\ to produce additional input to 
\cnn. 
While both \lstm\ \tvEmbs\ and \cnn\ \tvEmbs\ are region embeddings, their formulations 
are very different from each other; therefore, we expect that they complement each other and bring 
further performance improvements when combined.  
We will empirically confirm these conjectures in the experiments below.  
Note that being able to naturally combine several \tvEmbs\ is a strength of our framework, which 
uses unlabeled data to produce {\em additional input} to \lstm\ instead of pre-training.

\subsection{Semi-supervised experiments}
\label{sec:exp}


We used IMDB, Elec, and RCV1 for our semi-supervised experiments; 
20NG was excluded due to the absence of standard unlabeled data.  
Table \ref{tab:unlab} summarizes the unlabeled data. 
%
%
To experiment with \lstm\ \tvEmbs, 
we trained two \lstms\ (forward and backward) with 100 units each on unlabeled data. 
The training objective was 
to predict the next $k$ words 
where $k$ was set to 20 for RCV1 and 5 for others.  
Similar to \JZb, 
we minimized weighted square loss 
\newcommand{\lossWei}{\alpha}
$\sum_{i,j} \lossWei_{i,j}(\bz_i[j]-\bp_i[j])^2$ 
where $i$ goes through the time steps, 
$\bz$ represents the next $k$ words by a bow vector, 
and $\bp$ is the model output; 
$\lossWei_{i,j}$ were set to achieve negative sampling effect for speed-up; 
vocabulary control was performed 
for reducing undesirable relations between views, which 
sets the vocabulary of the target (i.e., the $k$ words) to the 30K most frequent words 
excluding function words (or stop words on RCV1). 
Other details followed the supervised experiments. 

Our semi-supervised one-hot bidirectional \lstm\ with pooling (\ohBiLstm) 
in row\#4 of Table \ref{tab:semi} used the two \lstm\ \tvEmbs\ 
trained on unlabeled data 
as described above, 
to produce additional input to 
one-hot \lstms\ in two directions (500 units each). 
Compared with the supervised \ohBiLstm\ (Table \ref{tab:sup}), 
clear performance improvements were obtained on all the datasets, 
thus, confirming the effectiveness of 
our approach. 

We review the semi-supervised performance of \wvLstms\ (Table \ref{tab:semi} row\#1). 
In \DLa\ 
the model consisted of a word embedding layer of 
512 dimensions, an \lstm\ layer with 1024 units, and 30 hidden units on top of the \lstm\ layer; 
the word embedding layer and the \lstm\ were pre-trained with unlabeled data and were fine-tuned with labeled data; 
pre-training used 
either the language model objective 
or autoencoder objective. 
The error rate on IMDB is from \DLa, and those on Elec and RCV1 are our best effort to 
perform pre-training with the language model objective.  
We used the same configuration on Elec as \DLa; 
however, on RCV1, which has 55 classes, 
30 hidden units turned out to be too few and we changed it to 1000. 
Although the pre-trained \wvLstm\ clearly outperformed the supervised \wvLstm\ (Table \ref{tab:sup}), 
it underperformed
the models with region \tvEmbs\ (Table \ref{tab:semi} row\#4,5).  

Previous studies on \lstm\ for text often convert words into pre-trained word vectors, 
and word2vec \cite{wvecNips13}
is a popular choice for this purpose. 
Therefore, we tested \wvBiLstm\ (word-vector bidirectional LSTM with pooling), 
whose only difference from \ohBiLstm\ is that the input to the \lstm\ layers is 
the pre-trained word vectors. 
The word vectors were optionally updated (fine-tuned) during training. 
Two types of word vectors were tested.  
The Google News word vectors were trained by word2vec on a huge (100 billion-word) news corpus 
and are provided publicly. 
On our tasks, \wvBiLstm\ using the Google News vectors (Table \ref{tab:semi} row\#2) performed relatively poorly. 
When word2vec was trained with the domain unlabeled data, better results were observed
after we scaled word vectors appropriately (Table \ref{tab:semi} row\#3). 
Still, it underperformed 
the models with region \tvEmbs\ (row \#4,5), 
which used the same domain unlabeled data. 
We attribute 
the superiority of the models with \tvEmbs\ 
to the fact that they learn, from unlabeled data, 
embeddings of {\em text regions}, which can convey higher-level concepts than single words in isolation. 

Now we review the performance of one-hot \cnn\ 
with one 200-dim \cnn\ \tvEmb\ (Table \ref{tab:semi} row\#5), 
which is comparable with
our \lstm\ 
with two 100-dim \lstm\ \tvEmbs\ (row\#4)
in terms of the dimensionality of \tvEmbs.  
The \lstm\ 
(row\#4) rivals or outperforms 
the \cnn\ 
(row\#5) on IMDB/Elec 
but underperforms it on RCV1.  Increasing the dimensionality of \lstm\ \tvEmbs\ from 
100 to 300 on RCV1, we obtain 8.62, but it still does not reach 7.97 of 
the \cnn. 
As discussed earlier, we attribute the superiority of 
one-hot \cnn\ 
on RCV1 
to its unique way of representing parts of documents via bow input.

\begin{table*}[t] 
\begin{center} 
\begin{tabular}{c|l|l|c|c|c|}
\cline{2-6} 
&         &\multicolumn{1}{|c|}{Unlabeled data usage} & IMDB & Elec & RCV1 \\
\cline{2-6} 
1 &      \ohBiLstm        & two \lstm\ \tvEmbAbb\              & 6.66 & 6.08 & 8.62 \\
\cline{2-6} 
2&\ohCnn\         \JZbq\ & $3\times$100-dim \cnn\ \tvEmbAbb\  & 6.51 & 6.27 & 7.71 \\
\hhline{~=====}
3&{\bf \ohBiLstm} &     $3\times$100-dim \cnn\ \tvEmbAbb\ &{\bf 5.94}& {\bf 5.55} & 8.52 \\
\cline{2-2} \cline{4-6}
4&\ohCnn         & \multicolumn{1}{|r|}{+ two {\bf \lstm\ \tvEmbAbb}} & 6.05 & 5.87 & {\bf 7.15} \\ 
\cline{2-6} 
\end{tabular} 
\caption{ \label{tab:combo}
Error rates (\%) obtained by combining \cnn\ \tvEmbAbb\ and \lstm\ \tvEmbAbb\ (rows 3--4).  
\lstm\ \tvEmbAbb\ were 100-dim each on IMDB and Elec, and 300-dim on RCV1.  
To see the combination effects, compare row\#3 with \#1, and compare row\#4 with \#2.  
}
\end{center} \end{table*}
\begin{table} 
\begin{center} 
\begin{tabular}{|l|c|c|c|c|}
\hline
                     & \multicolumn{1}{|c|}{{\small U}} &{\small IMDB} &{\small Elec}&{\small RCV1} \\
\hline
\ohCnn{\small+doc. \JZaq\ }    &{\small N}& 7.67 & 7.14 & --\\
Co-tr. {\em optimized} {\small \JZbq\ }&{\small Y}&{\small (8.06)}&{\small (7.63)}&{\small (8.73)} \\
Para.vector {\small [LM14]} &{\small Y}& 7.42 &  --  & -- \\
\wvLstm\ {\small \DLaq\ }    &{\small Y}& 7.24 & --   & --\\
\ohCnn{\small (semi.) \JZbq\ } &{\small Y}& 6.51 & 6.27 & 7.71 \\
\hline
{\bf Our best model} &{\small Y}&{\bf 5.94}&{\bf 5.55}&{\bf 7.15}\\
\hline
\end{tabular} 
\caption{ \label{tab:prev}
Comparison with previous best results.  Error rates (\%). 
``U'': Was unlabeled data used?    
``Co-tr. optimized'': co-training using \ohCnn\ as a base learner 
with parameters (e.g., when to stop) {\em optimized on the test data}; it demonstrates 
the difficulty of exploiting unlabeled data on these tasks. 
}
\end{center} \end{table}

\subsection{Experiments combining \cnn\ \tvEmbs\ and \lstm\ \tvEmbs}
\label{sec:comboexp}
In Section \ref{sec:combo} we noted that \lstm\ \tvEmbs\ and \cnn\ \tvEmbs\ 
can be naturally combined.  
We experimented with this idea in the following two settings.  

In one setting, 
\ohBiLstm\ takes additional input from five embeddings: two \lstm\ \tvEmbs\ used in Table \ref{tab:semi} 
and three \cnn\ \tvEmbs\ from \JZb\ 
obtained by three distinct combinations of 
training objectives and input representations, 
which are 
publicly provided.
These \cnn\ \tvEmbs\ were trained to be applied to text regions of size $k$ at every location 
taking bow input, 
where $k$ is 5 on IMDB/Elec and 20 on RCV1.  
We connect each of the \cnn\ \tvEmbs\ to an \lstm\ by aligning the centers 
of the regions of the former with the \lstm\ time steps; 
e.g., the \cnn\ \tvEmb\ result on 
the first five words is passed to the \lstm\ at the time step on the third word. 
In the second setting, 
we trained one-hot \cnn\ with these five types of \tvEmbs\ 
by replacing (\ref{eq:cnn}) 
$\cnnAct{\Wei\bx_{\iL} + \Bias}$ by 
$\cnnAct{\Wei\bx_{\iL} + \sum_j \Wtv{j}\xtv^j_{\iL} + \Bias}$ where 
$\xtv^j_{\iL}$ is the output of the $j$-th \tvEmb\ 
with the same alignment as above. 

Rows 3--4 of Table \ref{tab:combo} show the results of these two types of models.  
For comparison, we also show the results of 
the \lstm\ 
with \lstm\ \tvEmbs\ only 
(row\#1) and 
the \cnn\ 
with \cnn\ \tvEmbs\ only (row\#2). 
To see the effects of combination, compare row\#3 with row\#1, and compare row\#4 with row\#2.  
For example, 
adding the \cnn\ \tvEmbs\ to 
the \lstm\ 
of row\#1, the error rate on IMDB improved 
from 6.66 to 5.94, and 
adding the \lstm\ \tvEmbs\ to 
the \cnn\ 
of row\#2, the error rate on RCV1 improved 
from 7.71 to 7.15.  
The results indicate that, as expected, 
\lstm\ \tvEmbs\ and \cnn\ \tvEmbs\ complement each other and 
improve performance when combined.   

\subsection{Comparison with the previous best results} 
The previous best results in the literature are shown in Table \ref{tab:prev}.
More results of previous semi-supervised models can be found in \JZb, 
all of which clearly underperform the semi-supervised one-hot \cnn\ of Table \ref{tab:prev}.  
The best supervised results on IMDB/Elec of \JZa\ are in the first row, obtained by 
integrating a document embedding layer into one-hot \cnn.  
Many more of the previous results on IMDB can be found in \cite{LM14}, all of which are over 10\% 
except for 8.78 by bi-gram NBSVM \cite{WM12}.  
7.42 by paragraph vectors \cite{LM14} and 6.51 by \JZb\ were considered to be large improvements.  
As shown in the last row of Table \ref{tab:prev}, 
our new model further improved it to 5.94; 
also on Elec and RCV1, our best models exceeded the previous best results.  

\section{Conclusion}
\label{sec:conc}

Within the general framework of `region embedding + pooling' for text categorization, 
we explored region embeddings via one-hot \lstm. 
The region embedding of one-hot \lstm\ rivaled or outperformed that of the state-of-the art one-hot \cnn, proving its
effectiveness. 
We also found that the models with either one of these two types of region embedding 
strongly outperformed other methods including previous \lstm. 
The best results were obtained by combining the two types of region embedding 
trained on unlabeled data, 
suggesting that their strengths are complementary.  
As a result, we reported substantial improvements over the previous best results 
on benchmark datasets. 

At a high level, 
our results indicate the following.  
First, on this task, embeddings of text regions, which can convey higher-level concepts, are more useful than embeddings of single words in isolation.  
Second, useful region embeddings can be learned by working with one-hot vectors directly, 
either on labeled data or unlabeled data.  
Finally, a promising future direction might be to seek, under this framework, 
new region embedding methods with complementary benefits.  

\section*{Acknowledgements} 
We would like to thank anonymous reviewers for valuable feedback. 
This research was partially supported by NSF IIS-1250985, NSF IIS-1407939, and NIH R01AI116744.

\bibliography{lstm-tcat}
\bibliographystyle{icml2016}

\end{document}